\def\CLASSINPUTinnersidemargin{0.75in}
\def\CLASSINPUToutersidemargin{0.75in}
\def\CLASSINPUTbottomtextmargin{0.75in}
\documentclass[letterpaper, 10pt, conference]{IEEEtran}
\IEEEoverridecommandlockouts
\usepackage{cite}
\usepackage{float}
\usepackage{amsmath,amssymb,amsfonts}
\usepackage{algorithmic}
\usepackage{graphicx}
\usepackage{textcomp}
\usepackage{subcaption}
\usepackage{xcolor}
\usepackage{booktabs}
\usepackage{stfloats}     
\usepackage[hyphens]{url}
\usepackage[hidelinks]{hyperref}
\def\BibTeX{{\rm B\kern-.05em{\sc i\kern-.025em b}\kern-.08em
    T\kern-.1667em\lower.7ex\hbox{E}\kern-.125emX}}
\renewcommand{\IEEEtitletopspaceextra}{19pt}
\begin{document}

\title{Layered Risk Mapping for Autonomous Patient Transport in Expeditionary Medical Facilities}
\author{
    Lorena Maria Genua$^{1}$, 
    Sarvesh Prajapati$^{1}$, 
  Damla Leblebicioglu$^{1}$,
    Ta\c{s}k{\i}n Pad{\i}r$^{1\ddagger}$
    \thanks{$^{1}$Institute for Experiential Robotics, Northeastern    University, Boston, Massachusetts, USA.
    {\tt\small \{genua.l, prajapati.s, leblebicioglu.d, t.padir\}@northeastern.edu}}
    \thanks{$^\ddagger$Ta\c{s}k{\i}n Pad{\i}r holds concurrent appointments as a Professor of Electrical and Computer Engineering at Northeastern University and as an Amazon Scholar. This paper describes work performed at Northeastern University and is not associated with Amazon.}
    \thanks{This research was funded, in part, by the Advanced Research Projects Agency for Health (ARPA-H) Agreement No. 140D042590012. The views and conclusions contained in this document are those of the authors and should not be interpreted as representing the official policies, either expressed or implied, of the U.S. Government.}%
    \thanks{Project page: \textbf{\url{https://lrm-patient-transport.github.io/}}}
}

\maketitle
\begin{abstract}
In expeditionary medical facilities, routine patient transport imposes a compounding burden of personal protective equipment consumption, staff diversion, and elevated infection risk that becomes unsustainable under surge conditions. While autonomous wheelchairs could absorb this operational load, the safety-critical nature of patient transit within these highly unstructured and dynamic environments poses complex navigational challenges. To address this, we present a layered risk mapping framework that fuses four heterogeneous environmental hazards (terrain slope, static and dynamic obstacles, and semantic traversability) into a unified probabilistic cost surface via a Noisy-OR fusion model. In a paired Monte-Carlo evaluation, risk-informed fusion reduces collision rates from over 73\% to under 32\% and more than doubles obstacle clearance relative to a risk-unaware baseline. Additionaly, Noisy-OR achieves the highest clearance to obstacles and the lowest conditional peak risk across all tested hazard densities. We further validate the framework on a commercial powered wheelchair across three representative mission profiles in indoor and outdoor deployments, demonstrating that this architecture successfully meets the planning requirements of this previously unaddressed operational regime.
\end{abstract}


\section{Introduction}

\begin{figure}[h!]
    \centering
    \includegraphics[width=1.0\linewidth]{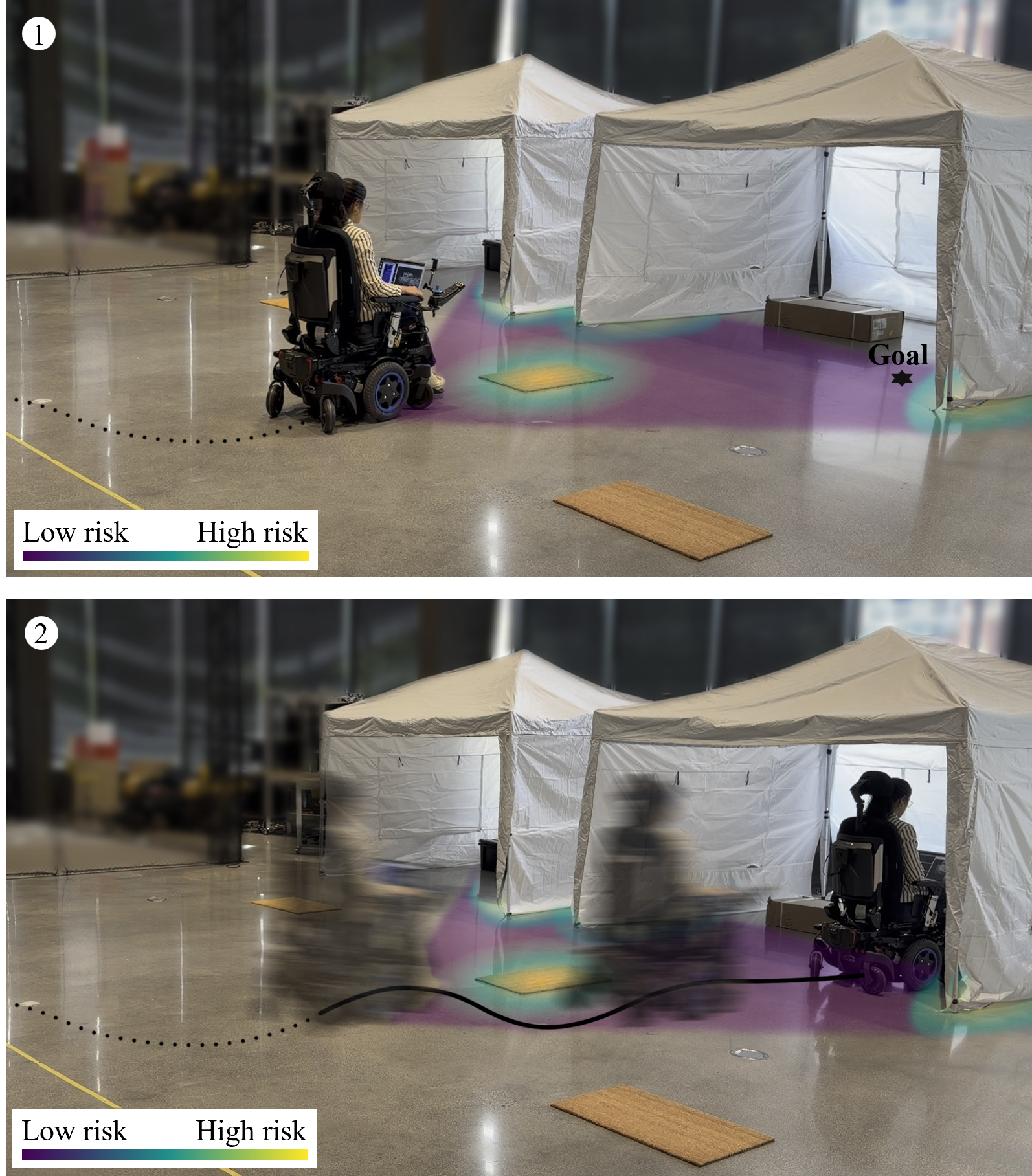} 
    \caption{An autonomous wheelchair transports patients between tented modules in an expeditionary medical facility, planning safe paths over a fused, multi-layer risk map of environmental hazards.}
    \label{fig:introduction}
    \vspace{-3ex}
\end{figure}

In May 2026, the World Health Organization declared the Bundibugyo Ebola outbreak in the Democratic Republic of the Congo and Uganda a Public Health Emergency of International Concern~\cite{who2026pheic}, the seventeenth Ebola outbreak in DRC alone, and renewed attention to the conditions under which clinical staff must operate within outbreak field hospitals.
In outbreak settings, each escorted patient transfer places clinical staff in repeated, direct contact with confirmed cases while simultaneously consuming a full set of personal protective equipment (PPE) and removing a trained clinician from bedside care.
During the 2022--2023 Sudan-EBOV outbreak in Uganda, healthcare workers accounted for 13.4\% of confirmed cases and 12.7\% of deaths, with PPE fatigue or misuse during direct patient-contact tasks identified as a primary transmission route~\cite{ref_ebola_logistics}.
Under surge conditions, when transfer frequency peaks and PPE stocks are already strained, this dual drain on personnel and infection-control resources becomes unsustainable.
An autonomous wheelchair fleet that transports patients between treatment modules without a human escort could absorb this burden entirely.
However, expeditionary medical facilities operate in an autonomy regime that neither indoor service robotics nor outdoor autonomous ground vehicles (AGVs) address: semi-structured gravel walkways and temporary boardwalks connecting tented treatment modules, surface transitions between earth, mud, and grass, dynamic obstacles in PPE-constrained corridors, and strict biohazard zone boundaries that must not be crossed.
\\This paper presents a layered risk mapping framework designed for this previously unaddressed operational regime.
Heterogeneous environmental hazards, including terrain slope, static and dynamic obstacles, and semantic traversability, are fused into a unified probabilistic cost surface via a Noisy-OR fusion model.
We evaluate the framework across three representative mission profiles: \emph{intake} (exterior triage to treatment module), \emph{intra-facility transfer} (module to module), and \emph{egress} (treatment area to discharge or evacuation zone).
These missions are validated both in simulation, comparing three risk fusion strategies across varying terrain and crowd configurations, and on a commercial
powered wheelchair in indoor and outdoor deployments.
Our results demonstrate that the framework successfully fulfills the planning requirements of this safety-critical domain, maintaining a stable obstacle clearance of 0.50\,m while reducing high-risk zone exposure by over 73\% compared to no-risk baseline.
Additionaly, among risk-informed methods, Noisy-OR achieved the highest clearance to obstacles and the lowest conditional peak risk across all tested hazard densities.
We also demonstrate the feasibility of our approach through real-world experiments in controlled indoor environment and unstructured outdoor terrains.
\\In summary, the primary contributions of this paper are:
\begin{itemize}
    \item We formalize autonomous patient transport in expeditionary medical facilities as a novel application domain.
    \item We propose a layered risk mapping framework that probabilistically fuses independent hazard layers into a unified cost surface using a Noisy-OR fusion model.
    \item We validate the framework through a paired Monte-Carlo study against three baseline fusion strategies and demonstrate its real-world feasibility on a commercial powered wheelchair across indoor and outdoor deployments.
\end{itemize}

\section{Related Work}
\label{sec:related_work}

\subsection{Outbreak Response and Expeditionary Medicine}

Deployable and expeditionary medical facilities, spanning military forward surgical teams, humanitarian field hospitals, and outbreak-response units present a navigational regime that existing robotics research has yet to address.
These environments combine rapidly erected modular infrastructure with heterogeneous surface transitions (packed gravel, temporary boardwalks, mud) and dynamic, PPE-constrained corridors that fundamentally differ from both structured indoor spaces and open outdoor terrain.
%
%
While robotic assistance in disaster-medicine contexts has focused primarily on search-and-rescue, casualty extraction~\cite{Watanobe2023, kumar2025swarm, roni-casualty-extraction} and medical deliveries~\cite{junior-drone-delivery, chen-medical-delivery}, autonomous patient transport has received comparatively little attention.
Platforms deployed for autonomous disinfection and contactless delivery in COVID-19 isolation wards demonstrated that robots can reduce staff exposure under surge conditions~\cite{ref_covid_robotics}, yet these systems operated strictly in indoor spaces with structured layouts.
In this paper, we demonstrate the application of an autonomous patient transport system designed to reduce clinical staff burden by reliably navigating the unstructured, hybrid terrains of deployable medical facilities without human escorts.

\subsection{Autonomous Wheelchairs and Assistive Patient Navigation}
Powered and semi-autonomous wheelchairs have been deployed primarily to restore mobility for individuals with motor, cognitive, or perceptual impairments in structured, indoor settings such as rehabilitation centers, care homes, and hospital corridors~\cite{ref_wheelchair_shared_control, ref_wheelchair_haptic_shared_ctrl}.
These platforms target use cases where a user retains partial intent but lacks reliable motor control, and their design has accordingly prioritized human-in-the-loop shared control architectures over full autonomy~\cite{ref_wheelchair_shared_control, ref_wheelchair_haptic_shared_ctrl, Xia_wheelchair_shared_2025, gupta-review-intelligent-wheelchair}. 
In parallel, autonomous patient transport has emerged~\cite{Grewal2018wheelchair, ref_wheelchair_hospital_transport}, motivated by the same staffing-efficiency arguments that drive service robot adoption~\cite{ref_smart_patient_robot}.
Baltazar et al.~\cite{ref_wheelchair_hospital_transport} demonstrated a fully autonomous wheelchair capable of accepting transport requests from a hospital information system, navigating across floors, and interfacing with elevators.
Navigation research for autonomous wheelchairs spans a spectrum of sensing and mapping strategies. 
Indoor systems predominantly rely on LiDAR and visual odometry to navigate within pre-mapped environments~\cite{Wang2020wheelchair, ref_wheelchair_indoor_outdoor}.
Outdoor wheelchair navigation remains far less explored; the limited existing work has addressed urban and structured outdoor environments through GPS, IMU, and LiDAR fusion~\cite{jianwei-outdoor, Spletzer2017outdoor, ref_wheelchair_indoor_outdoor}, but no approach considers traversability through heterogeneous or unstructured terrain.
Our work targets a fundamentally different operational regime: fully autonomous patient transport across the hybrid indoor-outdoor terrain of outbreak field hospitals, where surface heterogeneity and moving personnel and equipment must be reasoned over jointly; a combination unaddressed by any existing wheelchair navigation system.

\subsection{Risk-informed and Multi-Layer Risk Fusion}

Risk-informed navigation over unstructured terrain has been explored primarily for autonomous platforms that do not transport people.
RAMP~\cite{sharma2023ramp} fuses LiDAR-derived ground-plane risk into an Model Predictive Path Integral (MPPI) planner for off-road traversal, addressing a single geometric hazard source. 
Cai et al.~\cite{cai2023probabilistic} and Endo et al.~\cite{endo2023riskaware} probabilistically fuse terrain classification and slip-prediction models into a joint traversability distribution for off-road and planetary rovers.
Architectures that do combine multiple hazard sources rely on deterministic rules that treat layers as interchangeable magnitudes rather than distinct causal contributors: layered costmaps~\cite{lu2014layered} apply priority-based max-value overlay, allowing a single dominant layer to obscure risk from others, while driving-risk-field models for human-occupied vehicles sum weighted vehicle, road, and pedestrian potential fields~\cite{tan2024rcprf}, requiring hand-tuned weights with no probabilistic interpretation. 
A separate line of work fuses multi-modal hazard inputs end-to-end via learned cost networks~\cite{castro2023howdoesitfeel}, but such approaches require in-domain training data, unavailable for expeditionary medical facilities which are assembled anew at each deployment.
Our four hazard layers (terrain slope, static obstacles, dynamic obstacles, and semantic traversability) arise from physically and perceptually independent observations. 
Rather than tuning relative weights or letting one layer dominate, we combine each layer's independent failure probability into the joint probability that at least one violation occurs, a distinction that matters directly for patient safety, where an undetected or diluted hazard in any single layer translates to a real collision or rollover risk rather than a mere efficiency loss.
\begin{figure*}[t!]
    \centering
    \includegraphics[width=1.0\textwidth]{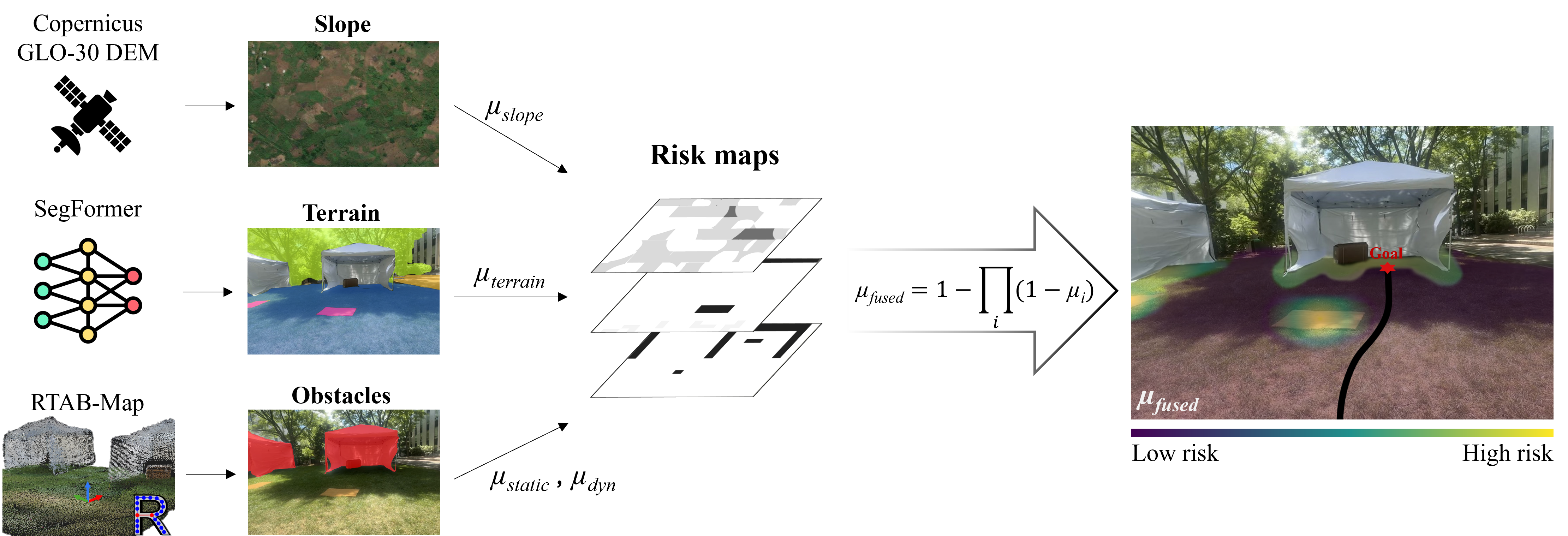} 
    \caption{\textbf{Overview of the multi-layer risk fusion architecture.} Global elevation data from Copernicus GLO-30 DEM yields the slope risk layer $\mathcal{R}_{slope}$, real-time visual streams are processed via a fine-tuned SegFormer model to resolve semantic terrain traversability $\mathcal{R}_{terr}$, and IMU-visual SLAM via RTAB-Map establishes static and dynamic obstacle boundaries ($\mathcal{R}_{static}, \mathcal{R}_{dyn}$). These heterogeneous inputs are normalized and fused using a Probabilistic Noisy-OR Fusion model to formulate a unified cost surface for safe local trajectory optimization.}
    \label{fig:methodology}
\end{figure*}
\section{Methodology}
Safe autonomous patient transport in expeditionary medical facilities requires reasoning jointly over terrain geometry, obstacles, and semantic context and failure modes that arise from distinct physical mechanisms and cannot be captured by a single sensor or cost function. We therefore represent each hazard source as a separate 2-D risk grid grounded in domain-specific criteria, and fuse the grids via a Noisy-OR model into a unified cost surface that drives an MPPI local planner in real time. The remainder of this section defines the four risk layers (Section~\ref{subsection:riskmap}) and the fusion and planning stack (Section~\ref{subsection:riskawarenav}).
\subsection{Risk Maps}
\label{subsection:riskmap}
Four hazard layers contribute to the risk map, indexed by $i \in \mathcal{L} = \{\text{slope}, \text{terr}, \text{static}, \text{dyn}\}$. Each layer is represented as a 2D grid map $\mathcal{R}_i : \mathbb{Z}^2 \to [0, 100]$, where $0$ denotes fully safe and $100$ denotes maximum risk. The layers capture terrain slope ($\mathcal{R}_{\text{slope}}$), semantic traversability ($\mathcal{R}_{\text{terr}}$), static obstacles ($\mathcal{R}_ {\text{static}}$), and dynamic obstacles ($\mathcal{R}_{\text{dyn}}$). Threshold values in the following subsections are calibrated to the mechanical envelope and regulatory limits of the Quickie Q500M mid-wheel-drive powered wheelchair (hereafter Q500M)~\cite{quickie_q500m_manual}, the platform used in our real-world experiments.

\vspace{0.6ex}
\textbf{Terrain slope} ($\mathcal{R}_{\text{slope}}$): Elevation data are fetched from FABDEM~\cite{hawker2022fabdem}, a global 30\,m bare-earth Digital Elevation Model derived from the Copernicus GLO-30 DEM with buildings and vegetation canopy removed, for a patch centered on the deployment site.
A slope map is computed from the DEM and thresholded into four risk tiers grounded in wheelchair accessibility regulations~\cite{ada2010} and the operational limits of Q500M power base upon which the LUCI safety payload is mounted.
Slopes below $5^\circ$ conform to the ADA maximum accessible ramp threshold of 1:12 ($\approx 4.8^\circ$)~\cite{ada2010} and are assigned zero risk.
Slopes between $5^\circ$ and $8^\circ$ exceed ADA public infrastructure guidelines but remain within the $8^\circ$ maximum safe operational slope specified by Sunrise Medical for the Q500M~\cite{quickie_q500m_manual}; increased motor load and lateral tracking deviation at these gradients~\cite{eder2023energy} warrant a moderate risk of 50.
Slopes between $8^\circ$ and $20^\circ$ exceed the manufacturer limit; significant wheel-slip and occupant center-of-gravity displacement are introduced at these gradients~\cite{wong2001theory}, and a high risk of 80 is assigned, remaining below the maximum risk reserved for slopes that exceed the platform's certified dynamic-stability envelope. 
This preserves finite (rather than lethal) traversal cost, allowing the planner to cross such terrain when no lower-risk route exists.
Slopes exceeding $20^\circ$ present an imminent rollover hazard well beyond the dynamic stability safety envelope of the platform, exceeding standard operational limits evaluated under ISO 7176-2 protocols~\cite{iso7176_2_2017} and are assigned the maximum risk of 100.

\textbf{Terrain traversability} ($\mathcal{R}_{\text{terr}}$): Real-time traversability is estimated from the RGB-D camera streams using a SegFormer-B2~\cite{xie2021segformer} fine-tuned on ADE20K~\cite{zhou2017ade20k}, which includes 150 semantic categories spanning outdoor classes (grass, road, water, wall, fence, building), providing the breadth of scene content required for wheelchair navigation.
The model outputs a per-pixel semantic label, which is then mapped to a traversability cost $c \in [0, 100]$ via a hand-crafted lookup table, where $0$ denotes freely traversable space and $100$ denotes a lethal obstacle.
Valid semantic pixels are back-projected into the 3D camera frame using sensor intrinsics and transformed into the fixed world frame via estimated camera extrinsics~\cite{hartley2004mvg} before being discretized into the 2D risk grid.
Costs are assigned in three tiers based on the mechanical interaction between the wheelchair's wheels and the terrain surface, a methodology consistent with the semantic traversability literature~\cite{sevastopoulos2022survey, larson2011offroad}.

\begin{itemize}
    \item \textit{Low risk} ($r \leq 10$): Hard, high-traction surfaces such as \emph{road}, \emph{path}, and \emph{floor} present minimal rolling resistance and negligible slip risk, assigned near-zero cost~\cite{wong2001theory, eder2023energy}.

    \item \textit{Intermediate risk} ($r = 20$--$65$): Deformable and
    high-resistance terrains such as \emph{grass}, \emph{sand}, \emph{mud}, and \emph{rock} are penalised progressively. Traversal over such surfaces induces wheel slip and increased energy consumption due to wheel-soil interaction~\cite{wong2001theory, cai2023probabilistic}. The ordering within this tier reflects increasing deformability and slip risk: grass and soil impose modest resistance, while mud and rock present significantly higher traversal difficulty~\cite{eder2023energy}.

    \item \textit{High risk} ($r = 100$): Structurally impassable or
    hazardous entities (e.g.\ \emph{wall}, \emph{fence}, \emph{water},
    \emph{building}) are assigned the maximum risk.
\end{itemize}

\textbf{Static obstacles} ($\mathcal{R}_{\text{static}}$): A static reference map is built prior to execution using RTAB-Map~\cite{labbe2019rtabmap}, which fuses RGB-D and inertial measurements to construct a globally consistent 3D point cloud of the environment. Points within the wheelchair's vertical clearance are extracted and discretized into a 2D occupancy grid, yielding $\mathcal{R}_{\text{static}}$, where cell values in $[0, 100]$ indicate unobserved or free space at $0$ and high-confidence obstacle returns at $100$.

\textbf{Dynamic obstacles} ($\mathcal{R}_{\text{dyn}}$): This layer is populated from the real-time mapping updates generated by RTAB-Map.
Moving hazards detected within the active mapping stream are projected into $\mathcal{R}_{\text{dyn}}$ as lethal costs.
\subsection{Risk-informed navigation}
\label{subsection:riskawarenav}
Classic navigation stacks typically combine costmaps using heuristic maximum-value blending or additive frameworks.
However, in unstructured and safety-critical environments, these approaches present severe limitations: maximum blending structurally dilutes compounding risks, while additive combinations can aggregate low-level sensor noise into artificial, impassable blockades.
To overcome this, we implement a \textit{Probabilistic Noisy-OR Fusion} framework that treats individual environmental hazards as independent causal factors capable of independently triggering a system safety violation.
\\Each risk map $\mathcal{R}_i : \mathbb{Z}^2 \to [0,100]$ is first normalized to a probabilistic risk index $\mu_i = \mathcal{R}_i / 100 \in [0,1]$, representing the independent probability that layer $i$ causes a navigational safety violation at a given grid cell.
Rather than utilizing standard additive or heuristic maximum-value layer blending strategies common in classic navigation stacks, which can dilute individual hazards, we implement a \textit{Probabilistic Noisy-OR Fusion} model. 
Each risk layer acts as an independent causal factor that can independently trigger a system safety failure. The fused risk index $\mu_{\text{fused}}$, representing the joint probability that at least one safety violation occurs, is formulated as:
\begin{equation}
    \mu_{\text{fused}} = 1 - \prod_{i}\left(1 - \mu_i\right)
    \label{eq:noisy_or_fused}
\end{equation}
The resulting cost surface drives trajectory optimization via MPPI control~\cite{williams2017mppi}, an information-theoretic MPC method that rolls out $K$ stochastic trajectory samples over a planning horizon of $H$ steps under a unicycle dynamics model and computes a cost-weighted update to the nominal control sequence. 
Each candidate rollout $k$ is scored as:
\begin{equation}
    J_k = \sum_{t=0}^{H-1} \mu_{\text{fused}}(x_t^k, y_t^k)
        + \lambda_{\text{term}} \lVert \mathbf{p}_H^k - \mathbf{p}_{\text{goal}} \rVert_2
        + \lambda_{\text{hdg}} (\tilde{\theta}_H^k)^2
\end{equation}
where $(x_t^k, y_t^k)$ is the predicted position of rollout $k$ at step $t$, $\mathbf{p}_H^k$ is the terminal position at the end of the horizon, $\mathbf{p}_{\text{goal}}$ is the target position, and $\tilde{\theta}_H^k$ is the wrapped heading alignment error to the target bearing. The weights $\lambda_{\text{term}}$ and $\lambda_{\text{hdg}}$ balance progress toward the goal against directional alignment.
\section{Experiments}
We evaluate the proposed framework in simulation and hardware.
To benchmark operational viability within an expeditionary field hospital layout, the autonomous wheelchair executes three distinct, representative patient transport missions:
\begin{itemize}
    \item \textbf{Intake scenarios}: Transporting incoming patients from an exterior triage or arrival zone into a designated treatment module.
    \item \textbf{Intra-facility scenarios}: Navigating between tented modules for diagnostics or specialized care.
    \item \textbf{Egress scenarios}: Routing patients from a treatment area to a discharge zone.
\end{itemize}
\subsection{Simulation}
\begin{figure}[b!]
    \centering
    \includegraphics[width=1.0\linewidth]{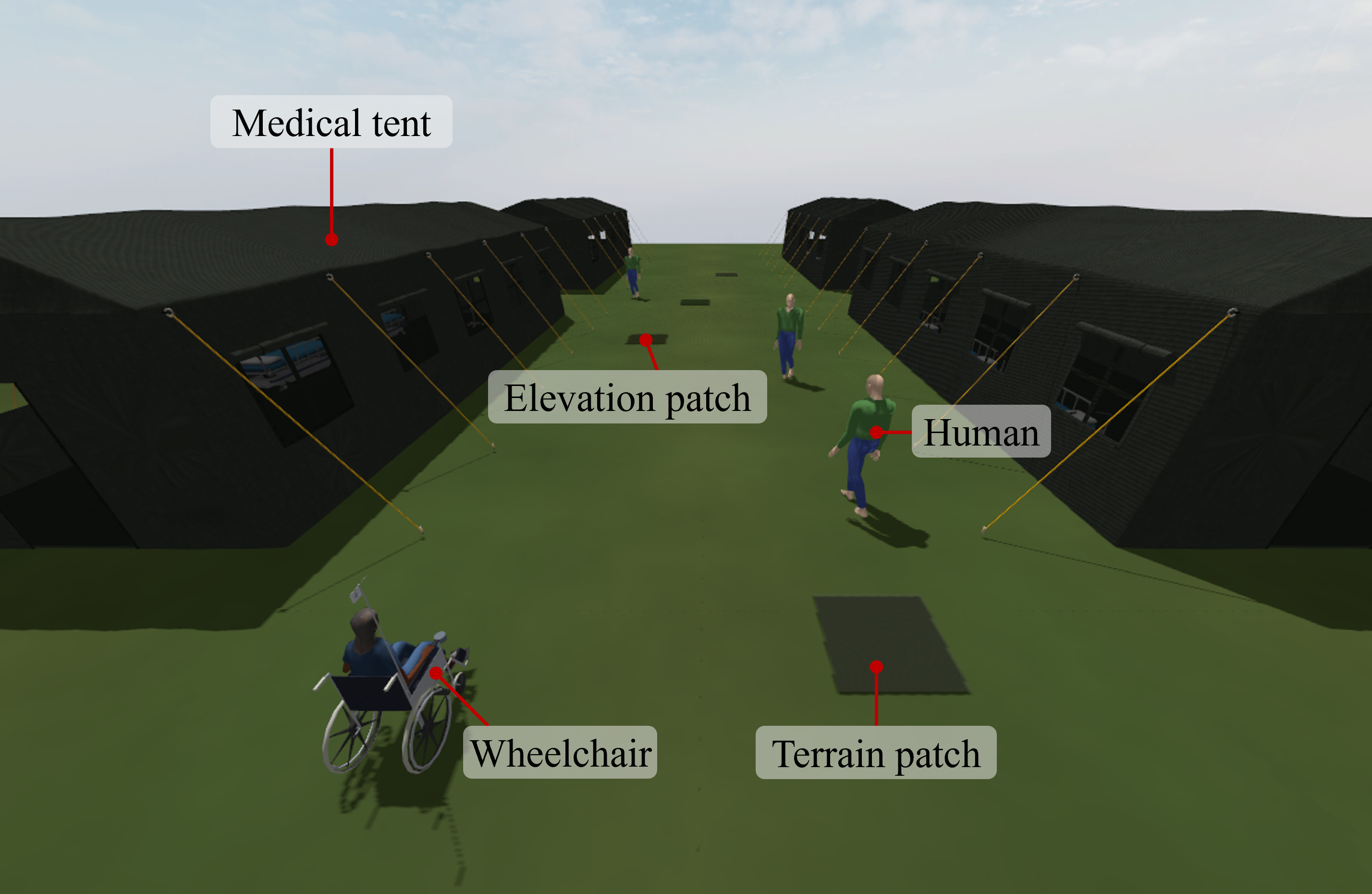} 
    \caption{\textbf{Simulation environment} used for the Monte-Carlo evaluation, modeling an expeditionary field hospital layout.}
    \label{fig:simulation}
\end{figure}
\begin{table*}[b!]
  \centering
  \caption{Navigation performance across risk-informed fusion methods at three environmental densities $\rho$. Values reported as mean $\pm$ std.}
  \label{tab:results}
  \footnotesize
  \begin{tabular*}{\textwidth}{l@{\extracolsep{\fill}}lcccc}
    \toprule
    Metric & $\rho$ & \textbf{No risk} & \textbf{Noisy-OR} & \textbf{Max} & \textbf{Log-odds} \\
    \midrule
    SPL
      & 1.0 & $0.77 \pm 0.18$ & $0.66 \pm 0.29$ & $0.71 \pm 0.26$ & $0.67 \pm 0.30$ \\
      & 1.5 & $0.77 \pm 0.18$ & $0.69 \pm 0.27$ & $0.69 \pm 0.28$ & $0.69 \pm 0.27$ \\
      & 2.0 & $0.77 \pm 0.18$ & $0.68 \pm 0.27$ & $0.65 \pm 0.31$ & $0.64 \pm 0.30$ \\
    \midrule
    Obstacle clearance (m)
      & 1.0 & $0.16 \pm 0.23$ & $\textbf{0.49} \pm 0.13$ & $0.42 \pm 0.15$ & $0.46 \pm 0.12$ \\
      & 1.5 & $0.14 \pm 0.22$ & $\textbf{0.50} \pm 0.07$ & $0.40 \pm 0.17$ & $0.44 \pm 0.12$ \\
      & 2.0 & $0.17 \pm 0.23$ & $\textbf{0.50} \pm 0.06$ & $0.42 \pm 0.14$ & $0.45 \pm 0.11$ \\
    \midrule
    Collision Rate (\%)
      & 1.0 & $73.3$ & $\textbf{17.8}$ & $22.2$ & $24.4$ \\
      & 1.5 & $82.2$ & $\textbf{24.4}$ & $31.1$ & $25.4$ \\
      & 2.0 & $75.6$ & $\textbf{17.8}$ & $22.2$ & $24.4$ \\
    \midrule
    Collisions
      & 1.0 & $3.94 \pm 1.50$ & $3.88 \pm 4.94$ & $\textbf{2.20} \pm 1.87$ & $2.27 \pm 2.33$ \\
      & 1.5 & $3.95 \pm 1.88$ & $\textbf{2.36} \pm 4.20$ & $5.43 \pm 11.06$ & $3.73 \pm 3.26$ \\
      & 2.0 & $4.00 \pm 1.86$ & $\textbf{1.50} \pm 0.53$ & $1.80 \pm 0.92$ & $2.00 \pm 1.55$ \\
    \midrule
    High-Risk Zone Entry Rate (\%)
      & 1.0 & $44.4$ & $8.9$ & $6.9$ & $\textbf{6.7}$ \\
      & 1.5 & $55.6$ & $13.3$ & $\textbf{8.9}$ & $11.1$ \\
      & 2.0 & $60.0$ & $13.3$ & $\textbf{11.1}$ & $\textbf{11.1}$ \\
    \midrule
    Conditional Peak Risk
      & 1.0 & $0.70 \pm 0.20$ & $\textbf{0.47} \pm 0.05$ & $0.69 \pm 0.20$ & $0.48 \pm 0.06$ \\
      & 1.5 & $0.73 \pm 0.19$ & $\textbf{0.50} \pm 0.16$ & $0.55 \pm 0.14$ & $0.52 \pm 0.16$ \\
      & 2.0 & $0.71 \pm 0.20$ & $\textbf{0.55} \pm 0.16$ & $0.56 \pm 0.23$ & $0.58 \pm 0.16$ \\
    \bottomrule
  \end{tabular*}
\end{table*}
We evaluate the proposed framework in a Gazebo simulation of a representative outbreak field hospital with four tented treatment modules arranged around a central staging area (Fig.~\ref{fig:simulation}).
Tent positions and the structural layout are fixed across all trials.
Heterogeneous terrain patches (gravel, mud, grass, rock) and elevation features are generated at the start of each trial using a configurable density parameter that scales the total patch count.
The environment also includes a simulated crowd of humans moving between tents along randomized trajectories per trial.
The simulated wheelchair mirrors the physical dimensions and kinematic constraints of a Q500M wheelchair.
\subsubsection{Experimental Procedure} We conduct a paired Monte-Carlo study evaluating four risk fusion methods across a representative expeditionary-facility setup.
Across these scenarios, we compare the following risk-fusion methods:
\begin{itemize}
    \item \textbf{Noisy-OR}: the formulation defined in Eq.~\eqref{eq:noisy_or_fused}.
    \item \textbf{Log-odds}: The Bayesian log-odds evidence accumulation defined as $\mu_{\text{fused}} = \sigma\!\left(\sum_i \ell_i\right)$ where $\ell_i$ represents the log-odds of layer $i$.
    \item \textbf{Maximum}: A conservative baseline utilizing standard max-value layer blending, $\mu_{\text{fused}} = \max_i \mu_i$, mirroring the default layer combination strategy implemented in the ROS 2 Navigation stack (Nav2).
    \item \textbf{No risk}: A risk-unaware baseline where the local planner sets $\mu_{\text{fused}} = 0$, bypassing the fused risk map to react solely to obstacles.
\end{itemize}
Each fusion method is evaluated over 50 runs per density level, with identical hazard configurations and crowd-trajectory seeds shared across methods within each run.
\subsubsection{Evaluation Metrics} For each run, we collect the following metrics.
\textit{Success-weighted Path Length} (SPL) measures path efficiency relative to the straight-line distance, conditioned on task success; failed runs contribute zero to this metric.
\textit{Obstacle clearance} (m) is the minimum robot-to-obstacle distance recorded over the course of a trial, capturing how safely the wheelchair navigates around static and dynamic obstacles.
\textit{Collision Rate} (\%) is the fraction of runs in which at least one collision occurs.
\textit{Collisions} counts the total number of contact events with static obstacles and humans per run.
\textit{High-Risk Zone Entry Rate} (\%) is the fraction of runs in which the wheelchair traverses at least one cell whose risk exceeds $0.3$.
\textit{Conditional Peak Risk} is the maximum risk value encountered during a run, conditioned on a risk zone entry; runs with no zone entry are excluded.
Each experiment is repeated across three hazard density levels ($\rho \in \{1.0, 1.5, 2.0\}$), which scale the number of terrain and slope patches generated at scene initialization.
At $\rho = 1.0$, the scene contains 24 terrain patches across four terrain types and 6 slope patches, for a total of 30 hazard patches.
At $\rho = 1.5$, patch counts increase to 32 terrain and 8 slope patches (40 total), and at $\rho = 2.0$ to 36 terrain and 9 slope patches (45 total), at which point patches are dense enough to cover a significant portion of the traversable area.
The number of humans in the scene is 4 and remains constant across all density levels.
\subsubsection{Results}
Table~\ref{tab:results} reports navigation performance across all fusion methods and density levels.
\\\textbf{Risk-informed methods vs.\ No Risk:}
Across all densities and all metrics every risk-informed fusion method outperforms the No Risk baseline.
The No Risk planner achieves the highest SPL ($0.77 \pm 0.18$ uniformly), since without terrain penalties it commits to direct routes without deviation.
However, this path efficiency comes at severe safety cost: collision rates reach $73.3\%$--$82.2\%$, mean obstacle clearance drops to $0.14$--$0.17$\,m, and high-risk zone entry rates climb to $44.4\%$--$60.0\%$ as density increases.
\\All risk-informed methods reduce collision rates to below $32\%$, maintain clearances above $0.40$\,m, and hold high-risk zone entry below $13.4\%$ at every density level.
The SPL reduction incurred by risk-informed planning is modest and consistent, reflecting deliberate route deviation around hazardous terrain.
\\\textbf{Comparison across risk-fusion methods:}
Noisy-OR is the most consistently safe method across all densities: it has the lowest obstacle clearance variance ($0.49$--$0.50\,$m, shrinking from $0.13$ to $0.06$ as density increases), the lowest collision rate, and the lowest conditional peak risk ($0.47$, $0.50$, $0.55$ for $\rho=1.0,1.5,2.0$), even though it has the \textit{highest} high-risk zone entry rate among the three risk-informed methods at every density (e.g., $8.9\%$ vs.\ $6.9\%$ for Max and $6.7\%$ for Log-odds at $\rho=1.0$).
This indicates that when Noisy-OR does route through a marginal-risk region, it keeps the peak severity of that exposure tightly bounded, and this stability holds regardless of hazard density.
Max, by contrast, shows a density-dependent blind spot that Noisy-OR does not: its conditional peak risk is highest at low density ($0.69$, nearly indistinguishable from the no-risk baseline's $0.70$) and falls to $0.55$--$0.56$ as density increases. This instability is not limited to peak risk: at $\rho=1.5$, Max's collision count exhibits a large variance ($5.43 \pm 11.06$), an order of magnitude higher than any other method or density in the table, pointing to occasional catastrophic failure runs beneath an otherwise unremarkable mean.
Log-odds shows a milder but opposite failure mode: its conditional peak risk climbs steadily as density increases ($0.48 \rightarrow 0.52 \rightarrow 0.58$), and it records the worst collision rate of the three risk-informed methods at every density ($24.4$--$25.4\%$).
\subsection{Field Validation}
\begin{figure*}[b!]
    \centering
    \begin{subfigure}[b]{0.48\textwidth}
        \centering
        \includegraphics[width=\textwidth]{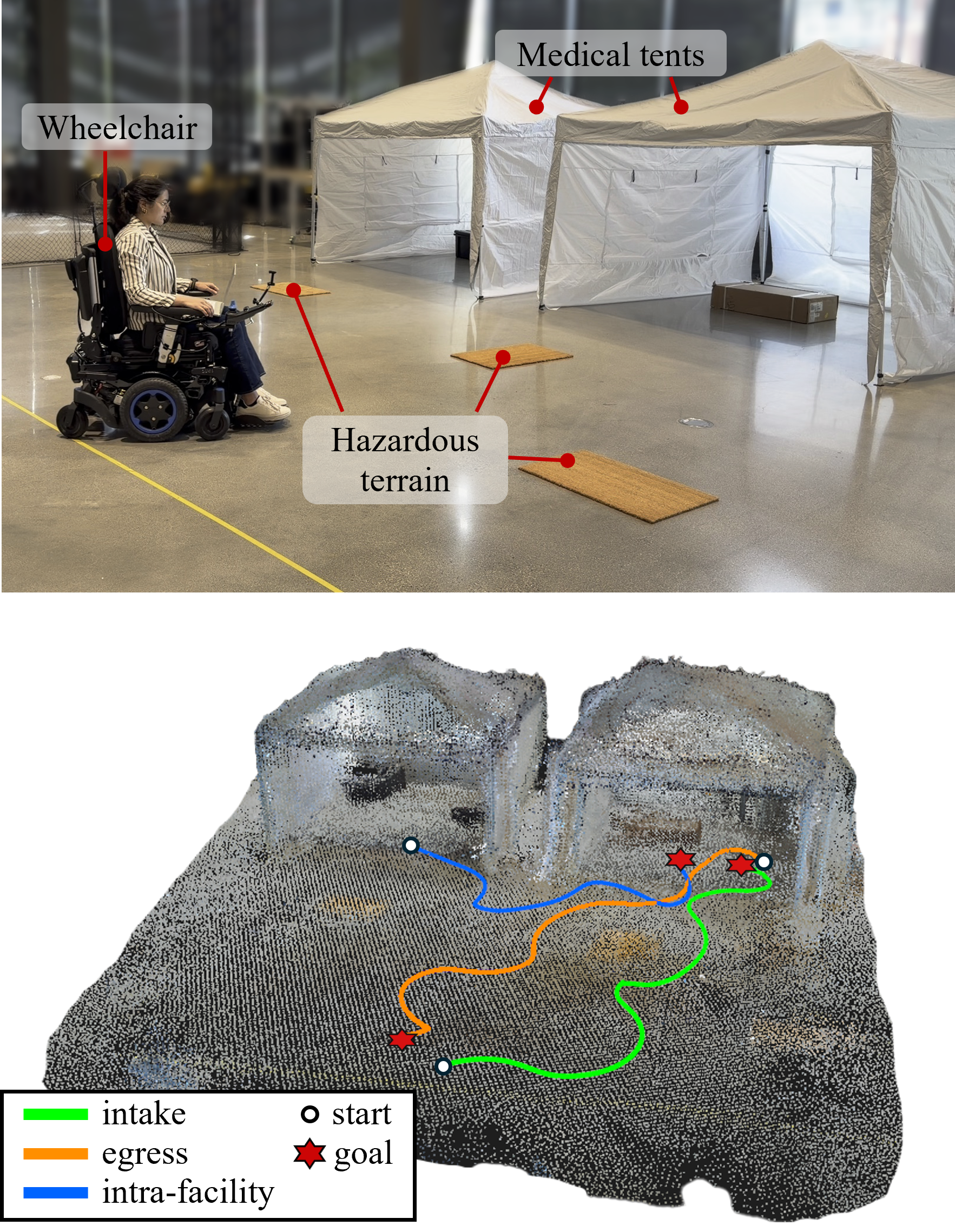}
        \caption{Indoor experiments}
        \label{fig:indoor}
    \end{subfigure}
    \hfill
    \begin{subfigure}[b]{0.48\textwidth}
        \centering
        \includegraphics[width=\textwidth]{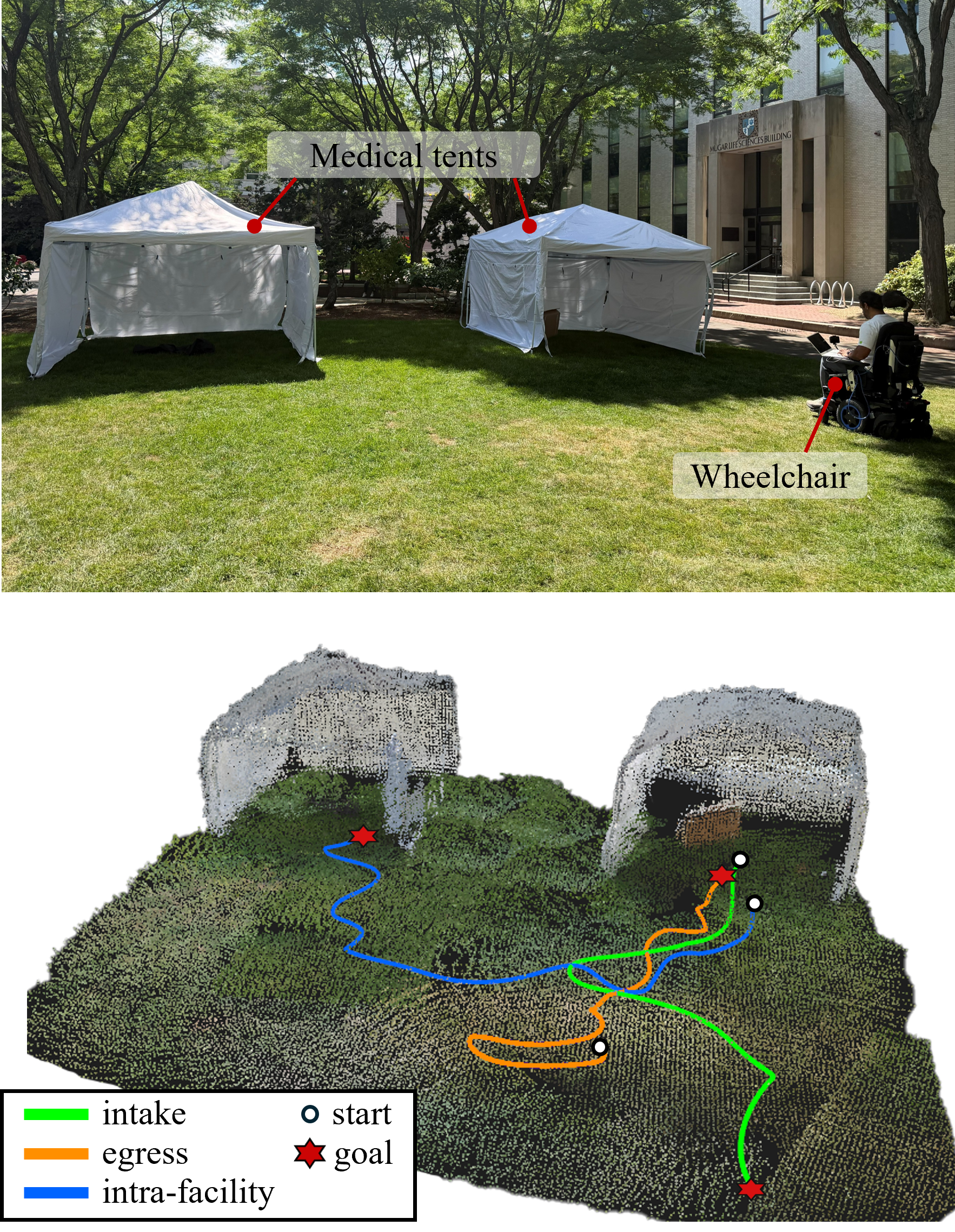}
        \caption{Outdoor experiments}
        \label{fig:outdoor}
    \end{subfigure}
    \caption{\textbf{Field validation.} Reconstructed 3D maps of executed trajectories for the three mission profiles: intake (green), egress (orange), and intra-facility transfer (blue). (a) Indoor validation with simulated terrain hazards. (b) Outdoor validation across unstructured grass terrain and elevation variation.}
    \label{fig:hardware_experiments}
\end{figure*}
\begin{figure}[h!]
    \centering
    \includegraphics[width=\linewidth]{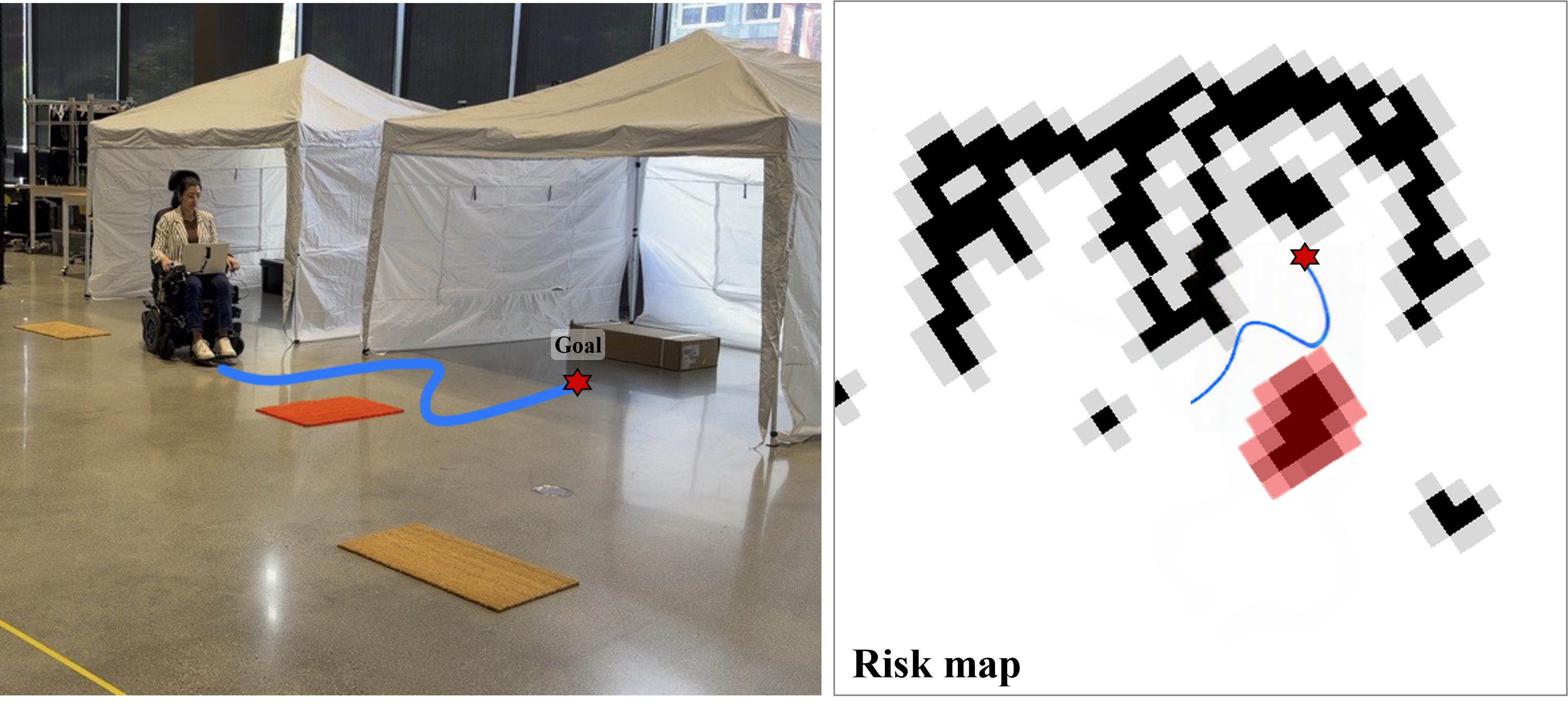} 
    \caption{Mid-transit snapshot (left) demonstrating real-world path execution alongside its corresponding localized risk map surface (right). The hazardous terrain patch on the floor produces a high-risk region (red) in the map, forcing the MPPI planner to dynamically route around them.}
    \label{fig:hardware-exp-risk-map}
\end{figure}

For real-world validation, we utilize a commercial Quickie Q500M mid-wheel-drive powered wheelchair equipped with 
\begin{figure}[H]
    \centering
    \includegraphics[width=0.9\linewidth]{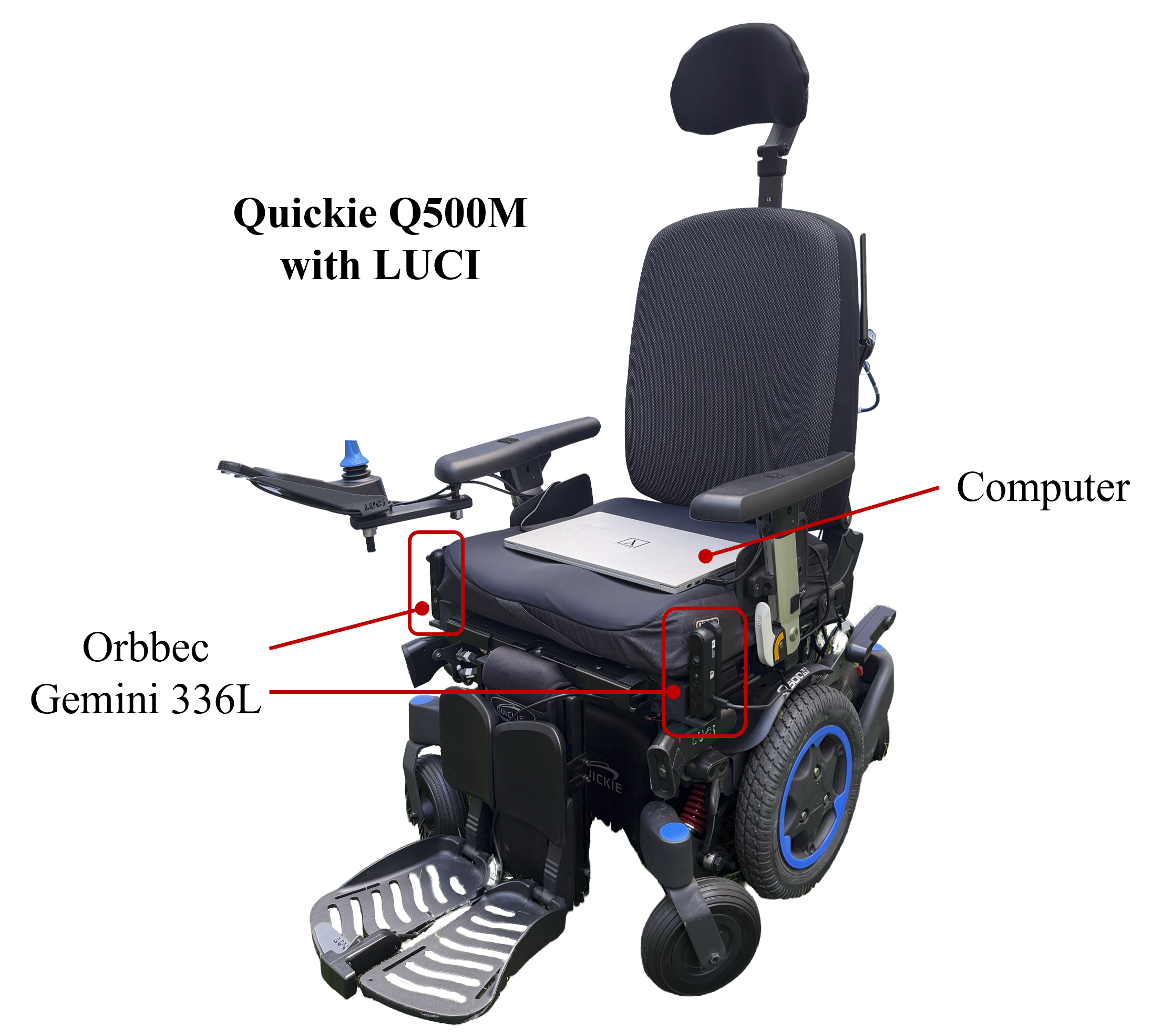} 
    \caption{The commercial Quickie Q500M powered wheelchair with the LUCI safety payload used for real-world validation.}
    \label{fig:wheelchair_hardware}
    \vspace{0ex}
\end{figure}
\noindent a LUCI safety payload (Fig.~\ref{fig:wheelchair_hardware}).
The wheelchair has two forward-facing Orbbec Gemini 336L RGB-D cameras mounted on the seat rails.
All computation is handled by an onboard Razer x Lambda Tensorbook laptop (Intel Core i7-11800H CPU; NVIDIA GeForce RTX 3080 Mobile GPU, 16 GB VRAM; 64 GB RAM; Ubuntu 22.04 LTS).
All risk grids use a cell size of 0.30~m over a 10~m local window centered on the wheelchair. The semantic traversability and dynamic-obstacle layers refresh at 15~Hz from the RGB-D stream; the static and slope layers are precomputed offline. The fused cost surface is recomputed at each MPPI planning cycle. MPPI uses $K = 1000$ rollouts over a horizon of $H = 40$ steps at $\Delta t = 0.1$\,s, with control-noise standard deviations $\sigma_v = 0.2$ m/s and $\sigma_\omega = 0.1$ rad/s and temperature $\tau = 0.4$. The weights $\lambda_{\text{term}} = 12.0$ and $\lambda_{\text{hdg}} = 0.5$ were tuned on outdoor trials and held fixed across all reported experiments. We treat the normalized cost $\mu_i = \mathcal{R}_i / 100$ as an uncalibrated pseudo-probability; layer-specific probability calibration is future work. The full ADE20K class-to-cost lookup is provided on the project page.
We conduct trials in two physical settings that mirror the indoor-outdoor terrain heterogeneity of expeditionary facilities.
The indoor setting reproduces the tented treatment-module layout within a controlled facility space, with a simulated hazardous floor patch placed between modules (Fig.~\ref{fig:hardware_experiments}, left).
The outdoor setting deploys the same tent configuration on unstructured grass terrain with natural elevation variation and transitions between modules (Fig.~\ref{fig:outdoor}).
In both settings, the wheelchair executes the same three mission profiles evaluated in simulation: intake, intra-facility transfer, and egress.
The starting and ending positions were chosen randomly.
Fig.~\ref{fig:hardware_experiments} shows the RTAB-Map reconstructed 3D point clouds with executed trajectories overlaid for each mission profile, indoors (a) and outdoors (b).
For the indoor trials, the Intake profile completed ($10.0\text{m}$) in $65.94\text{s}$, the Intra-facility transfer ($6.23\text{m}$) in $42.62\text{s}$, and the Egress profile ($8.59\text{m}$) in $47.08\text{s}$.
In the unstructured outdoor terrain completed the Intake profile ($6.93\text{m}$) in $45.27\text{s}$, the Intra-facility transfer ($11.12\text{m}$) in $69.13\text{s}$, and the Egress profile ($9.09\text{m}$) in $54.53\text{s}$.
Fig.~\ref{fig:hardware-exp-risk-map} illustrates the local risk map generated during one representative trial: the hazardous floor patch is identified as a high-risk region (red), causing the planner to deviate and route around it before re-converging on the goal.

\noindent Crucially, this happens because the fused risk map is updated locally and continuously as the robot navigates. Hazards are assessed reactively within the sliding local map window, allowing the MPPI planner to dynamically adjust to unfolding environmental features.
\\The field validation also highlighted some limitations. 
Currently, all components execute onboard a laptop, which introduces strict resource competition when simultaneously running high-frequency semantic segmentation (SegFormer), local planning, and spatial mapping. 
Future deployments will transition to a dedicated embedded computing hardware solution (e.g., NVIDIA Jetson AGX Orin) to fully decouple sensor processing from wheelchair control loops. 
Furthermore, the primary edge cases stemmed mainly from localized tracking degradation within visual SLAM (RTAB-Map).
Especially during the outdoor experiments, in feature-sparse environments and during sudden lighting transitions, visual odometry drift occurred, which affected the accuracy of the local planning. 
While the underlying visual SLAM subsystem does not perform perfectly under all operational conditions, resolving tracking drift is outside the scope of this work. In future iterations, we will integrate a different SLAM method better suited for feature-sparse, unstructured environments.

\section{Conclusion}
This paper presented a multi-layer risk mapping framework for autonomous patient transport in expeditionary medical facilities, an operational regime combining hybrid indoor-outdoor terrain that existing indoor service robots and outdoor autonomous ground vehicles do not jointly address.
The framework fuses four heterogeneous hazard layers, terrain slope, static obstacles, dynamic obstacles, and semantic traversability, into a unified cost surface via a Noisy-OR fusion model, which drives an MPPI-based local planner in real time.
In simulation, all risk-informed fusion methods delivered safety improvements over a risk-unaware baseline across three mission profiles and three environmental hazard densities, reducing collision rates from over $73\%$ to below $32\%$ and high-risk zone entry from over $44\%$ to below $13.4\%$.
Among the risk-informed methods evaluated, the Noisy-OR formulation minimized conditional peak risk while maintaining the most consistent obstacle clearance.
Field validation on a commercial Q500M powered wheelchair, conducted across indoor and outdoor layouts, confirmed that the framework transfers to real hardware and reliably routes around physical hazards while completing all three mission profiles.

\bibliographystyle{IEEEtran}
\bibliography{references}

@inproceedings{xie2021segformer,
  title     = {{SegFormer}: Simple and Efficient Design for Semantic
               Segmentation with Transformers},
  author    = {Xie, Enze and Wang, Wenhai and Yu, Zhiding and
               Anandkumar, Anima and Alvarez, Jose M. and Luo, Ping},
  booktitle = {Advances in Neural Information Processing Systems (NeurIPS)},
  year      = {2021}
}

@inproceedings{zhou2017ade20k,
  title     = {Scene Parsing Through {ADE20K} Dataset},
  author    = {Zhou, Bolei and Zhao, Hang and Puig, Xavier and
               Fidler, Sanja and Barriuso, Adela and Torralba, Antonio},
  booktitle = {Proceedings of the IEEE Conference on Computer Vision
               and Pattern Recognition (CVPR)},
  year      = {2017}
}

@book{hartley2004mvg,
  title     = {Multiple View Geometry in Computer Vision},
  author    = {Hartley, Richard and Zisserman, Andrew},
  edition   = {2nd},
  publisher = {Cambridge University Press},
  year      = {2004}
}

@article{sevastopoulos2022survey,
  title   = {A Survey of Traversability Estimation for Mobile Robots},
  author  = {Sevastopoulos, Christos and Konstantopoulos, Stasinos},
  journal = {IEEE Access},
  volume  = {10},
  pages   = {96331--96347},
  year    = {2022}
}

@article{tan2024rcprf,
  author  = {Tan, Yiyang and others},
  title   = {{RCP-RF}: A Comprehensive Road-Car-Pedestrian Risk Management Framework based on Driving Risk Potential Field},
  journal = {IET Intelligent Transport Systems},
  year    = {2024},
  doi     = {10.1049/itr2.12508}
}

@inproceedings{castro2023howdoesitfeel,
  author    = {Guaman Castro, Mateo and Triest, Samuel and Wang, Wenshan and Gregory, Jason M. and Sanchez, Felix and Rogers III, John G. and Scherer, Sebastian},
  title     = {How Does It Feel? Self-Supervised Costmap Learning for Off-Road Vehicle Traversability},
  booktitle = {2023 IEEE International Conference on Robotics and Automation (ICRA)},
  pages     = {931--938},
  year      = {2023},
  publisher = {IEEE}
}

@inproceedings{larson2011offroad,
  title     = {Off-Road Terrain Traversability Analysis and Hazard
               Avoidance for {UGV}s},
  author    = {Larson, Jacoby and Trivedi, Mohan and Bruch, Michael},
  booktitle = {IEEE Intelligent Vehicles Symposium},
  year      = {2011}
}

@article{Spletzer2017outdoor,
  author    = {Spletzer, John and others},
  title     = {A smart wheelchair ecosystem for autonomous navigation in urban environments},
  journal   = {Autonomous Robots},
  volume    = {41},
  number    = {3},
  pages     = {519--538},
  year      = {2017},
  month     = {March},
  publisher = {Springer US},
  doi       = {10.1007/s10514-016-9549-1},
  issn      = {1573-7527}
}

@misc{who2026pheic,
  author       = {{World Health Organization}},
  title        = {Epidemic of {Ebola} Disease caused by {Bundibugyo} virus in the {Democratic Republic of the Congo} and {Uganda} determined a public health emergency of international concern},
  year         = {2026},
  month        = may,
  day          = {17},
  howpublished = {\url{https://www.who.int/news/item/17-05-2026-epidemic-of-ebola-disease-in-the-democratic-republic-of-the-congo-and-uganda-determined-a-public-health-emergency-of-international-concern}},
  note         = {Accessed: Jun. 2026}
}

@book{wong2001theory,
  title     = {Theory of Ground Vehicles},
  author    = {Wong, J. Y.},
  edition   = {3rd},
  publisher = {John Wiley \& Sons},
  year      = {2001}
}

@inproceedings{cai2023probabilistic,
  title     = {Probabilistic Traversability Model for Risk-Aware Motion
               Planning in Off-Road Environments},
  author    = {Cai, Xiaoyi and Everett, Michael and Sharma, Lakshay and
               Osteen, Philip R. and How, Jonathan P.},
  booktitle = {IEEE/RSJ International Conference on Intelligent Robots
               and Systems (IROS)},
  year      = {2023}
}

@inproceedings{sharma2023ramp,
  author    = {Sharma, Lakshay and Everett, Michael and Lee, Dexter and Cai, Xuesu and Osteen, Philip and How, Jonathan P.},
  title     = {{RAMP}: A Risk-Aware Mapping and Planning Pipeline for Fast Off-Road Ground Robot Navigation},
  booktitle = {2023 IEEE International Conference on Robotics and Automation (ICRA)},
  pages     = {5730--5736},
  year      = {2023},
  publisher = {IEEE}
}

@inproceedings{endo2023riskaware,
  author    = {Endo, Masafumi and Taniai, Tatsunori and Yonetani, Ryo and Ishigami, Genya},
  title     = {Risk-Aware Path Planning via Probabilistic Fusion of Traversability Prediction for Planetary Rovers on Heterogeneous Terrains},
  booktitle = {2023 IEEE International Conference on Robotics and Automation (ICRA)},
  year      = {2023},
  publisher = {IEEE}
}

@inproceedings{lu2014layered,
  author    = {Lu, David V. and Hershberger, Dave and Smart, William D.},
  title     = {Layered Costmaps for Context-Sensitive Navigation},
  booktitle = {2014 IEEE/RSJ International Conference on Intelligent Robots and Systems (IROS)},
  pages     = {709--715},
  year      = {2014},
  publisher = {IEEE}
}

@inproceedings{eder2023energy,
  title     = {Predicting Energy Consumption and Traversal Time of Ground
               Robots for Outdoor Navigation on Multiple Types of Terrain},
  author    = {Eder, Matthias and Steinbauer-Wagner, Gerald},
  booktitle = {IEEE/RSJ International Conference on Intelligent Robots
               and Systems (IROS)},
  year      = {2023}
}

@article{labbe2019rtabmap,
  title   = {{RTAB-Map} as an Open-Source Lidar and Visual {SLAM} Library
             for Large-Scale and Long-Term Online Operation},
  author  = {Labb\'e, Mathieu and Michaud, Fran\c{c}ois},
  journal = {Journal of Field Robotics},
  volume  = {36},
  number  = {2},
  pages   = {416--446},
  year    = {2019}
}

@misc{ada2010,
  author       = {{U.S. Department of Justice}},
  title        = {Information and Technical Assistance on the {A}mericans with {D}isabilities {A}ct},
  howpublished = {\url{https://www.ada.gov/law-and-regs/}},
}

@manual{quickie_q500m_manual,
  title        = {{QUICKIE} {Q500} Series Power Wheelchair Owner's Manual},
  author       = {{Sunrise Medical}},
  year         = {2024},
  note         = {Rev. B}
}

@techreport{iso7176_2_2017,
  type        = {Standard},
  author      = {},
  key         = {ISO 7176-2:2017},
  year        = {2017},
  title       = {Wheelchairs — Part 2: Determination of dynamic stability of electric wheelchairs},
  institution = {International Organization for Standardization}
}

@article{ref_ebola_logistics,
  author    = {Namusobya, Zainab and others},
  title     = {Preparedness of healthcare workers for the {Ebola} outbreak
               in {Mubende} and {Kassanda} districts, {Uganda}},
  journal   = {PLOS ONE},
  year      = {2024},
  url       = {https://pmc.ncbi.nlm.nih.gov/articles/PMC12421453/}
}

@Article{junior-drone-delivery,
AUTHOR = {Fagundes-Júnior, Leonardo A. and Barcelos, Celso O. and Silvatti, Amanda Piaia and Brandão, Alexandre S.},
TITLE = {UAV–UGV Formation for Delivery Missions: A Practical Case Study},
JOURNAL = {Drones},
VOLUME = {9},
YEAR = {2025},
NUMBER = {1},
ARTICLE-NUMBER = {48},
URL = {https://www.mdpi.com/2504-446X/9/1/48},
ISSN = {2504-446X},
DOI = {10.3390/drones9010048}
}

@article{ref_covid_robotics,
  author    = {Tamantini, Christian and {Scotto di Luzio}, Francesco
               and Cordella, Francesca and Pascarella, Giuseppe
               and Agr{\`o}, Felice Eugenio and Zollo, Loredana},
  title     = {A Robotic Health-Care Assistant for {COVID-19} Emergency:
               A Proposed Solution for Logistics and Disinfection
               in a Hospital Environment},
  journal   = {{IEEE} Robotics \& Automation Magazine},
  volume    = {28},
  number    = {1},
  pages     = {71--81},
  year      = {2021},
  doi       = {10.1109/MRA.2020.3044953}
}

@INPROCEEDINGS{williams2017mppi,
  author={Williams, Grady and Wagener, Nolan and Goldfain, Brian and Drews, Paul and Rehg, James M. and Boots, Byron and Theodorou, Evangelos A.},
  booktitle={2017 IEEE International Conference on Robotics and Automation (ICRA)}, 
  title={Information theoretic MPC for model-based reinforcement learning}, 
  year={2017},
  volume={},
  number={},
  pages={1714-1721},
  doi={10.1109/ICRA.2017.7989202}}

@inproceedings{kumar2025swarm,
  author    ={Kumar, T. Rajesh and Nandhini, T. J. and Jumaniyazova, Ilmira and Abdulla, H. and Jumaniyozov, Yunus},
  title     ={Swarm Robotics for Search and Rescue Operations in Disaster Zones Using Particle Swarm Optimization ({PSO}) Algorithms},
  booktitle ={Proceedings of the 2025 International Conference on Networks and Cryptology (NETCRYPT)},
  pages     ={870--875},
  year      ={2025},
  publisher ={IEEE},
  doi       ={10.1109/NETCRYPT63559.2025.11102388}
}

@article{hawker2022fabdem,
  author  = {Hawker, Laurence and Uhe, Peter and Paulo, Luntadila and Sosa, Jeison
             and Savage, James and Sampson, Christopher and Neal, Jeffrey},
  title   = {A 30\,m global map of elevation with forests and buildings removed},
  journal = {Environmental Research Letters},
  year    = {2022},
  volume  = {17},
  number  = {2},
  pages   = {024016},
  doi     = {10.1088/1748-9326/ac4f D5},
}

@article{Watanobe2023,
  title={Disaster Rescue via Multi-Robot Collaboration: Development, Control, and Deployment},
  author={Yutaka Watanobe and Raihan Kabir and Ryuma Aoba and Ayato Ohashi and Shunsuke Ogata and Mizuki Shiga and Kota Tsuruno and Tsuyoshi Anazawa and Keitaro Naruse},
  journal={Journal of Robotics and Mechatronics},
  volume={35},
  number={1},
  pages={85-98},
  year={2023},
  doi={10.20965/jrm.2023.p0085}
}

@article{ref_wheelchair_hospital_transport,
  author    = {Baltazar, Andr{\'e} R. and Petry, Marcelo R.
               and Silva, Manuel F. and Moreira, Ant{\'o}nio Paulo},
  title     = {Autonomous Wheelchair for Patient's Transportation
               on Healthcare Institutions},
  journal   = {{SN} Applied Sciences},
  volume    = {3},
  number    = {3},
  pages     = {313},
  year      = {2021},
  doi       = {10.1007/s42452-021-04304-1}
}

@article{Wang2020wheelchair,
  author    = {Wang, Chaoqun and Xia, Min and Meng, Max Q.-H.},
  title     = {Stable Autonomous Robotic Wheelchair Navigation in the Environment With Slope Way},
  journal   = {IEEE Transactions on Vehicular Technology},
  volume    = {69},
  number    = {10},
  year      = {2020},
  publisher = {IEEE},
  doi       = {10.1109/TVT.9143493}
}

@inproceedings{Grewal2018Wheelchair,
  author    = {Grewal, H. S. and Jayaprakash, N. T. and Matthews, A. and Shrivastav, C. and George, K.},
  title     = {Autonomous Wheelchair Navigation in Unmapped Indoor Environments},
  booktitle = {2018 IEEE International Instrumentation and Measurement Technology Conference (I2MTC)},
  address   = {Houston, TX, USA},
  month     = may,
  year      = {2018},
  pages     = {1--6},
  doi       = {10.1109/I2MTC.2018.8409854},
  publisher = {IEEE}
}

@article{ref_smart_patient_robot,
  author    = {Kim, Bumsoo and Hyun, Jaeho and Yang, Bomi
               and Moon, Youngjin and Choi, Jaesoon},
  title     = {Development of Smart Patient Care Robot with Enhanced
               Autonomous Navigation Through Path Optimization
               in Hospital Wards},
  journal   = {Scientific Reports},
  volume    = {16},
  year      = {2026},
  doi       = {10.1038/s41598-026-36664-2}
}

@article{ref_wheelchair_shared_control,
  author    = {Yokota, Sho and Hashimoto, Hiroshi and Ohyama, Yasuhiro
               and She, Jin-Hua},
  title     = {Shared Control of an Electric Wheelchair Considering
               Physical Functions and Driving Motivation},
  journal   = {{IEEE} Transactions on Neural Systems and
               Rehabilitation Engineering},
  year      = {2020},
  doi       = {10.1109/TNSRE.2020.3003969},
  url       = {https://pmc.ncbi.nlm.nih.gov/articles/PMC7432419/}
}

@article{ref_wheelchair_haptic_shared_ctrl,
  author    = {Vargas, Francisco J. and others},
  title     = {Model-Based Shared Control Approach for a Power
               Wheelchair Driving Assistance System Using a
               Force Feedback Joystick},
  journal   = {Frontiers in Control Engineering},
  volume    = {4},
  pages     = {1058802},
  year      = {2023},
  doi       = {10.3389/fcteg.2023.1058802}
}

@article{roni-casualty-extraction,
author = {Saputra, Roni and Alattar, Ahmad and Rijanto, Estiko and Taqriban, Rilo and Septevani, Athanasia and Rakicevic, Nemanja and Kormushev, Petar},
year = {2025},
month = {01},
pages = {1-1},
title = {Rescue Robots for Casualty Extraction: A Comprehensive Review},
volume = {PP},
journal = {IEEE Access},
doi = {10.1109/ACCESS.2025.3637567}
}

@Article{jianwei-outdoor,
AUTHOR = {Cui, Jianwei and Yu, Siji and Shang, Yucheng and Dai, Yuxiang and Zhang, Wenyi},
TITLE = {Research on Outdoor Navigation of Intelligent Wheelchair Based on a Novel Layered Cost Map},
JOURNAL = {Actuators},
VOLUME = {14},
YEAR = {2025},
NUMBER = {2},
ARTICLE-NUMBER = {46},
URL = {https://www.mdpi.com/2076-0825/14/2/46},
ISSN = {2076-0825},
DOI = {10.3390/act14020046}
}

@inproceedings{ref_wheelchair_indoor_outdoor,
author = {Habha, Loiy and Trivedi, Urvish and Alqasemi, Redwan and Dubey, Rajiv},
title = {Autonomous Wheelchair Indoor-Outdoor Navigation System through Accessible Routes},
year = {2021},
isbn = {9781450387927},
publisher = {Association for Computing Machinery},
doi = {10.1145/3453892.3453997},
booktitle = {Proceedings of the 14th PErvasive Technologies Related to Assistive Environments Conference},
pages = {199–202},
numpages = {4},
location = {Corfu, Greece},
series = {PETRA '21}
}

@article{gupta-review-intelligent-wheelchair,
author={Atulan,Gupta and Chowdhury,Kanan R. and Nusrat,Farheen and Schoen,Marco P.},
year={2025},
title={Evolution and Emerging Trends in Intelligent Wheelchair Control: A Comprehensive Review},
journal={Machines},
volume={14},
number={1},
pages={33},
url={http://ezproxy.wpi.edu/login?url=https://www.proquest.com/scholarly-journals/evolution-emerging-trends-intelligent-wheelchair/docview/3297448991/se-2},
}

@article{Xia_wheelchair_shared_2025,
doi = {10.1088/1741-2552/ae2ccc},
url = {https://doi.org/10.1088/1741-2552/ae2ccc},
year = {2025},
month = {dec},
publisher = {IOP Publishing},
volume = {22},
number = {6},
pages = {066037},
author = {Xia, Yuchen and Wei, Yuxuan and Li, Songwei and Mai, Ximing and Luo, Ruijie and Zhu, Xiangyang and Meng, Jianjun},
title = {A potential field shared control approach for wheelchair navigation via brain–computer interface},
journal = {Journal of Neural Engineering}}

@article{chen-medical-delivery,
author = {Chen, Kaiyuan and Zhao, Wanpeng and Liu, Yongxi and Wang, Na and Xia, Yuanqing and Liang, Wannian and Wang, Shuo},
title = {UAV–UGV Cooperative Trajectory Optimization and Task Allocation for Medical Rescue Tasks in Post-Disaster Environments},
journal = {Unmanned Systems},
volume = {0},
number = {0},
pages = {1-15},
year = {0},
doi = {10.1142/S2301385027500452},
URL = {https://doi.org/10.1142/S2301385027500452},
}

\end{document}